\documentclass{llncs}

\usepackage{times}
\usepackage[latin1]{inputenc}
\usepackage[T1]{fontenc}
\usepackage{lscape}

\usepackage[english]{babel}%
\usepackage[pdftex]{graphicx}
\usepackage{amssymb} 
\usepackage[ruled,vlined,titlenumbered]{algorithm2e}

\newtheorem{pte}{Property}

\SetKw{nil}{nil}
\SetFuncSty{\textit}

\SetVline

\SetKwFor{Foreach}{for each}{do}{endfor}

\dontprintsemicolon

\SetKwIf{Debut}{EDebut}{}{}{}{}

\title{Parallel Strategies Selection}
\author{Anthony Palmieri$^{1}$ and Jean-Charles R{\'e}gin$^{2}$ and Pierre Schaus$^{3}$}
\institute{$^{1}$ Huawei Technologie France\\ 
$^{2}$ Univ. Nice Sophia Antipolis, CNRS, I3S, UMR 7271, 06900 Sophia Antipolis, France\\
$^{3}$Univ. Louvain-La-Neuven, Belgique\\
  \email{anthony.palmieri@hotmail.fr, jcregin@gmail.com, pierre.schaus@uclouvain.be}}

\begin{document}

\maketitle

\begin{abstract}
We consider the problem of selecting the best variable-value strategy for solving a given problem in constraint programming. We show that the recent Embarrassingly Parallel Search method (EPS) can be used for this purpose. EPS proposes to solve a problem by decomposing it in a lot of subproblems and to give them on-demand to workers which run in parallel. Our method uses a part of these subproblems as a simple sample as defined in statistics for comparing some strategies in order to select the most promising one that will be used for solving the remaining subproblems. For each subproblem of the sample, the parallelism helps us to control the running time of the strategies because it gives us the possibility to introduce timeouts by stopping a strategy when it requires more than twice the time of the best one. Thus, we can deal with the great disparity in solving times for the strategies. 
The selections we made are based on the Wilcoxon signed rank tests because no assumption has to be made on the distribution of the solving times and because these tests can deal with the censored data that we obtain after introducing timeouts. 
The experiments we performed on a set of classical benchmarks for satisfaction and optimization problems show that our method obtain good performance by selecting almost all the time the best variable-value strategy and by almost never choosing a variable-value strategy which is dramatically slower than the best one. Our method also outperforms the portfolio approach consisting in running some strategies in parallel and is competitive with the multi armed bandit framework.
\end{abstract}

\section{Introduction}

In absence of specific knowledge, defining an efficient variable-value strategy for guiding the search for solutions of a given problem  is not an easy task. Thus some generic variable-value strategies have been defined. They either try to apply generic principles like the first fail principle (i.e., we should try to fail as quickly as possible) \cite{haralick80} or try to detect underlined relations between variables and constraints. In the first case, we have strategies like min-domain which selects the variable having the minimum domain size, max-constrained which prefers variables involved in a lot of constraints, or min-regret which selects the variable which may lead to the largest increase in the cost if it is not selected. The latter case is mainly formed by the impact based strategy \cite{refalo04}, weighted degree strategy \cite{boussemart04b} and the activity based strategy \cite{michel12}. However, selecting a priori the best variable-value strategy is not an easy task, because
there is no strategy better than another in general and because it is quite difficult to identify the types of problems for which a strategy is going to perform well. In addition, there is no robustness among the strategies. Any variable-value strategy can give good results for a problem and really bad results for some others. It is not rare to see ratio of performance for a pair of strategy going to 1 to 20  (and even more sometimes) according to the problems which are solved. 

Unfortunately, there is almost no way to compare the performance of variable-value  strategies on a problem without solving it. Since strategies explore in different ways the search space  and since their pruning performance are not regular it is difficult to compare their behavior before the end of the resolution. 

So, selecting the right strategy is not an easy task and selecting the wrong strategy may be time consuming.

Our problem can also be seen as the automatic selection of the most efficient algorithm among a predefined set of algorithms, for solving a given problem \cite{rice76,kotthoff12,kotthoff16}. Usually two types of approaches are considered \cite{gagliolo06}. Either we try to determine statically, that is a priori, which is the best algorithm or we dynamically compute the best algorithm to use for each step of the problem solving. Both cases use a set of instances of the problem from which they learn different criteria that will be used to take a decision. 

We propose an original approach which is not based on machine learning but on the statistical estimation of the best algorithm.  Our approach does not require to deal with a set of instances and use some sampling technique that are usually more accurate. It exploits the decomposition proposed by the embarrassingly parallel search (EPS) method recently developed \cite{regin13,regin14}.

EPS proposes to solve a problem by decomposing it into a large number of subproblems consistent with the propagation (i.e., there is no immediate failure triggered by the initial propagation of a subproblem). We propose to use a part of these subproblems for comparing the strategies in order to select the most promising one for solving the whole problem.
Instead of comparing the strategies after solving the whole problem, we compare the strategies for each subproblem of the sample. This opens the door to fast methods for detecting the best strategy. We measure the solving time for each subproblem and each strategy and we eliminate the strategies that are statistically proved to be less efficient by a Wilcoxon signed rank test. 
At the end, either only one strategy remains or a set of non distinguishable strategies will remain. In this latter we select the one having the smallest mean.  

Since for each subproblem the solving times for the strategies may strongly vary, it is necessary to add a mechanism to control the time spent in the strategy selection and to stop some computations after a given amount of time. In other words we introduce timeouts. From a statistical point of view, this means that we may have censored data. 
We show that by defining appropriately these timeouts the results of the Wilcoxon signed rank test remains valid if timeouts were not considered.
Solving each subproblem with each strategy in parallel allows us to define relative timeouts: we stop a strategy when it requires more than twice the solving time of the best strategy.

It is important to note that our method does not require to know the distribution of the solving times (we made some experiments showing that the distributions vary according to the problems or to the strategy, and there is no general guidelines). 

Our method can be distinguished from the machine learning approaches in two ways:
\begin{itemize}
\item The relation between the data from which we take our decision and the instance to solve is stronger in our case because we consider subproblems of the instance and not some other instances.
\item We do not try to learn any criteria and we do no try to estimate any solving time. We just want to select the strategy that is the most promising one for the given instance. The comparison is relative and valid only for the given instance. Our results are statistically validated.
\end{itemize}
 
 The paper is organized as follows. First we show the principles of our method on an example. Then,  we recall some preliminaries. Next, we detail the different steps of our approach. 
 We present some related work and some experiments on a set of benchmarks, for which we compare our results with classical portfolio and a multi armed bandit method. At last we conclude.
  
\section{Selection Principles}

We propose to present the principles of our method on a didactic example obtained from the all-interval series, a common benchmark.

Our method proceeds by elimination of strategies until there is only one remaining. 

We consider 4 strategies ($S_1,S_2,S_3,S_4$). The initial problem has been decomposed into 300 subproblems from which we randomly select only 10 subproblems for the sake of clarity.

We could consider each subproblem in turn and run all the strategies on it in parallel. The drawback of this approach is that the running times are not regular and that some strategies may perform very badly for some subproblems compared to other strategies. For instance, here are the runtimes (in milliseconds) for each subproblem:

 \begin{center}
  \begin{tabular}{  | *{5}{c|}}
    \hline

subproblem & $S_1$ & $S_2$ & $S_3$ & $S_4$ \\
 \hline
1 & 62 & 408 & 80 & 150 \\
 \hline
2 & 90 & 1 134 & 92 & 154\\
 \hline
3 & 155 & 1 904 & 158 & 233\\
 \hline
4 & 231 & 1 451 & 250 & 407\\
 \hline
5 & 198 & 1 580 & 197 & 422\\
 \hline
6 & 146 & 803 & 170 & 144\\
 \hline
7 & 62 & 611 & 54 & 115\\
 \hline
8 & 63 & 389 & 111 & 86\\
 \hline
9 & 167 & 560 & 163 & 670\\
 \hline
10 & 83 & 736 & 120 & 232\\
\hline
$\Sigma$ & 1 257 & 9 576 & 1 395 & 2 613\\
 \hline
  \end{tabular}
\end{center}

With this approach the total time for selecting the best strategy is  $1257+9576+1395+2613=14841$, that is more than $10$ times the best runtime. Since there are $300$ subproblems to solve and since we selected $10$, then we can expect a total solving time  around $30$ times the runtime of the best strategy for our $10$ subproblems that is $1.26 \times  30=  37.8s$\footnote{We do not claim that this computation is accurate. We present it only for understanding the intuitive idea.}. This means that the time allocated to the strategy selection, named selection time,  may require more than 40\% of the solving time. 
In practice running all the strategies on each subproblem in the sample might take up to 90\% of the solving time that would be taken by the best strategy to solve all the subproblems. 
Our objective is to keep the overhead induced by the selection strategy minimal. Therefore some timeouts must be introduced (measured with respect to the best strategy on each subproblem). Timeouts may cause censored measures that must be carefully treated by statistical methods. 

We propose to deal with censored data and proceed by steps. 
\begin{enumerate}
\item For each subproblem we compare the strategies, but we introduce a timeout limit for each computation corresponding to 2 times the runtime obtained by the best strategy. 
\item We select the strategy having the smallest total time (timeouts are counted as their values). If this strategy was stopped by a timeout for some subproblems we run it again on these subproblems without timeouts. We repeat this step until the strategy, denoted by $s_b$ having the smallest total time without timeout has been selected. 
\item We compare all the strategies against $s_b$ by using the Wilcoxon signed rank test.
All strategies significantly slower than $s_b$ are eliminated. If $s_b$ is rejected by the Wilcoxon test against $s_x$ (in theory this can happens even if $s_b$ has a better mean) then $s_b$ is eliminated and replaced by $s_x$. Note that this latter case never happens in the 10,000s of tests we made. 
\item Eventually, if some strategies cannot be distinguished by the Wilcoxon signed rank test then we select the strategy performing the best on the sample.
\end{enumerate}
Note that in any case we have a strong statistical support of our choice.

With timeouts corresponding to twice the runtime of the best strategy for each subproblem we obtain the following table:
 \begin{center}
  \begin{tabular}{  | *{6}{c|}}
    \hline
 
\# & timeout & $S_1$ & $S_2$ & $S_3$ & $S_4$ \\
 \hline
1 & $2 \times 62= 124 $ & 62 & TO & 80 & TO \\
 \hline
2 & $2 \times 90= 180$ & 90 & TO & 92 & 154\\
 \hline
3 & $2 \times 155=310$ & 155 & TO & 158 & 233\\
 \hline
4 & $2 \times 231=462$& 231 & TO & 250 & 407\\
 \hline
5 & $2 \times 197=394$& 198 & TO & 197 & TO\\
 \hline
6 & $2 \times 144=288$& 146 & TO & 170 & 144\\
 \hline
7 & $2 \times 54=108$ & 62 & TO & 54 & TO\\
 \hline
8 & $2 \times 63=126$ & 63 & TO & 111 & 86\\
 \hline
9 & $2 \times 163=326$ & 167 & TO & 163 & TO\\
 \hline
10 & $2 \times 83=166$ & 83 & TO & 120 & TO\\
\hline
$\Sigma$ & & 1 257 & 2 484 & 1 395 & 2 142\\
 \hline
  \end{tabular}
\end{center}

It is important to remark that the best strategy for the whole problem is not the best one for each subproblem. In practice it happens frequently that the best strategy has some timeouts.

The Wilcoxon signed rank test considers the difference in response within pairs. Then it ranks the absolute values of these differences.  
The sum $W^+$ of the ranks for the positive difference is the \emph{Wilcoxon signed rank statistic} and has mean
      $\mu_{W^+}=\frac{n(n+1)}{4}$. 
      The Wilcoxon signed rank test rejects the hypothesis that there is no systematic differences within pairs when the rank sum $W^+$ is far from its mean.
      
Consider we want to compare the strategies $S_1$ and $S_3$. For each subproblem we compute the difference $time(S_1) - time(S_3)$. Then, we rank the absolute values of these differences and we add a sign in front of these ranks corresponding of the signs of the differences. For instance, for the first subproblem we have  $time(S_1) - time(S_3)=62-80=-16$, $16$ is the $6^{th}$ values so its rank is $6$. The sign rank is $-6$ because the difference is negative. Then, we compute $W^+$, the sum of the positive ranks. The following table
shows that we have $W^+=1+5+4=10$.
 \begin{center}
  \begin{tabular}{  | *{5}{c|}}
    \hline
sub problem & $S_1$ & $S_3$ & $S_1 - S_3$ & signed rank \\
 \hline
1 & 62 &  80 &  -18 & -6\\
 \hline
2 & 90 & 92 & -2 & -2\\
 \hline
3 & 155  & 158 & -3 & -3\\
 \hline
4 & 231  & 250 & -19 & -7\\
 \hline
5 & 198  & 197 & 1 & +1\\
 \hline
6 & 146  & 170 & -24 & -8\\
 \hline
7 & 62  & 54 & 8 & +5\\
 \hline
8 & 63  & 111 & -48 & -10\\
 \hline
9 & 167  & 163 & 4 & +4\\
 \hline
10 & 83  & 120 & -37 & -9\\
 \hline
  \end{tabular}
\end{center}
    
        We consider a one-tailed test ($S_3=S_1$ or $S_3 > S_1$) with a significance level of $0.05$.
      
      The critical value of $W$ for $N=10$ at $p \leq 0.05$ is $10$. Therefore the result is significant and we can conclude that $S_1$ is better than $S_3$. So, we can eliminate $S_3$.
      
      We repeat this process between $S_1$ and the other strategies. We will prove that we can perform the calculations by using the timeouts values if these values are defined by any value greater than twice the maximum positive difference because in this case the positive ranks will not change for any value greater than this timeout. 
      For instance, when we compare $S_1$ and $S_4$ there is only one positive difference equal to $2$ (for subproblem $6$, we have $146-144=2$), so for each subproblem $j$ we can set the timeout to any value $v$ such that $v > time(S_1,j)$ and $|time(S_1,j)-v| > 2$, because this will not impact the rank of value $2$ and so the value of $W^+$.
      
       If we apply this process for our example, the comparison against $S_1$ will eliminate all the other strategies. 

In conclusion, $S_1$ is selected. This leads to a resolution time of about $39.4$s.

   \section{Backgrounds}
   
   \subsection{Statistics}
   
   These definitions are due to  \cite{Moore2009}.
   \subsubsection{Simple random samples.}
   
   A \emph{simple random sample (SRS)} of size $n$ consists of $n$ individuals from the population chosen in such a way that every set of $n$ individuals has an equal chance to be the sample actually selected. We select an SRS by labeling all the individuals in the population and selecting randomly a sample of the desired size. Notice that an SRS not only gives each individual an equal chance to be chosen (thus avoiding bias in the choice) but gives every possible sample an equal chance to be chosen.

%
%
      \subsubsection{Wilcoxon Signed Rank Test for Matched Pairs.}
      
      Our data do not respect a Normal distribution and timeouts are introduced leading to right censored data. Thus, we cannot use common method like t distribution for comparing the  strategies and nonparametric tests have to be considered. Bootstrap method and permutation test are based on the idea of applying the method many times which will be too much time consuming in our case. In addition, these methods require heavy computing which is not acceptable for our application.  Hence we use the Wilcoxon Signed Rank Test.
      
Since we aim at comparing the performance of two algorithms we consider \emph{matched pairs} design, which compares just two observations. The idea is that matched subjects are more similar than unmatched subjects so that comparing responses within a number of pairs is more efficient then comparing the responses of groups of randomly assigned subjects. 
Matched pairs data are analyzed by taking the difference within the matched pairs to produce a single sample. The one sample statistic is applied on this difference data in order to compare the matched pairs data. 
      
The \emph{Wilcoxon signed rank test} (WSR test) for matched pairs  is defined as follows.      
      Draw an SRS of size $n$ from a population for a matched pairs study and take the difference in responses within pairs. Rank the absolute values of these differences. The sum $W^+$ of the ranks for the positive difference is the \emph{Wilcoxon signed rank statistic}. If the distribution of the responses is not affected by the different treatments within pairs, then $W^+$ has mean
      $\mu_{W^+}=\frac{n(n+1)}{4}$ and standard deviation  $\sigma_{W^+}=\sqrt{\frac{n(n+1)(2n+1)}{24}}$. 
      Difference of zero are discarded before ranking. Ties among the absolute differences are handled by assigning average ranks. 
      
      The WSR test rejects the hypothesis that there is no systematic differences within pairs when the rank sum $W^+$ is far from its mean.
      
P-values (i.e., the probability computed assuming that null hypothesis is true, that the test statistic will take a value at least as extreme as that actually observed) for the signed rank test are based on the sampling distribution of $W^+$ when the null hypothesis is true. 
      P-values can be computed from  the exact distribution (from software or tables) or obtained from a Normal approximation with continuity correction.

   \subsection{Embarrassingly Parallel Search}
   
   This presentation comes from \cite{regin14}.
   
   The idea of the Embarassingly Parallel Search (EPS) is to decompose statically the initial problem into a huge number of subproblems that are consistent with the propagation (i.e., running the propagation mechanism on them does not detect any inconsistency). These subproblems are added to a queue which is managed by a master. Then, each idle worker takes a  subproblem from the queue and solves it. The process is repeated until all the subproblems have been solved. 
   
The decomposition is made by selecting a set $V$ of $k$ variables and then by searching all instantiations of $V$ that are consistent with the propagation. There is  no specific variable-value strategy used to find these instantiations. The number of generated subproblems depends on the size of $V$ which is determine by successive computations.

The assignment of the subproblems to workers is dynamic and there is no communication between the workers. EPS is based on the idea that if there is a large number of subproblems to solve then the resolution times of the workers will be balanced even if the resolution times of the subproblems are not.
In other words, load balancing is automatically obtained in a statistical sense.
Interestingly, experiments of \cite{regin13} have shown that the number of subproblems does not depend on the initial problem but rather on the number of workers. Moreover, they have shown that a good decomposition has to generate more than $30$ subproblems per worker.

\section{Method}

\subsection{Simple random sample}

We use EPS to decompose the initial problem into a huge set of subproblems. Thus the population is the set of these subproblems. 
The SRS is built by selecting randomly $k$ sup-problems from the set of subproblems. 

Since we do not want to spend a lot of time in the time allocated to the strategy selection, the sample should not contain more than  1\% of subproblems. If $k=30$ subproblems seems to be the minimum number of subproblem to consider, then we need to have at least $3,000$ subproblems.

\subsection{Comparison of Strategies}

Strategies are compared by using the WSR test on the SRS previously defined.

For each subproblem of the SRS we run the strategies in parallel and we stop the slowest ones when they require twice the time of the best strategy.
Then, we select the strategy having the smallest sum of solving times for all the subproblems of the SRS. If this strategy was stopped by a timeout for some subproblems we run it again on these subproblems without timeout. We repeat this step until the strategy, denoted by $S_b$, having the smallest total time without timeout has been selected. 
Next,  we compare all the strategies against $S_b$ by using the WSR test performed on some modified data. All strategies significantly slower than $S_b$ are eliminated. 
If at a moment, the strategy $S_b$ is rejected by the Wilcoxon test against another strategy $S_x$, then  timeouts are removed for $S_x$ and we  use a t-test for deciding whether $S_x$ should become the best strategy. In this latter case we simply replace $S_b$ by $S_x$. 

In any we have a strong statistical support of our selection. 

Our Hypotheses are\\
$H_0$: there is no difference between data of both Strategies. \\
$H_a$: scores are systematically higher for the second Strategy.

In order to make sure that the result of the test remains valid when exact solving times are considered instead of timeout values, we proceed as follows.
Consider that we compare $S_b$ and $S_i$. Let us show that if we set for each subproblem $j$ the timeout to a value $to(j) > d_{bi}^{\max} + time(S_b,j)$ where $d_{bi}^{\max}$ the largest positive value of $time(S_b) - time(S_i)$ for all the subproblems of the SRS then the test is valid if exact solving times are considered instead of timeouts.


\begin{pte}\label{d12max}
Let $d_{bi}^{\max}$ the largest positive value of $time(S_b,j) - time(S_i,j)$, and \\
$rank(d_{bi}^{\max})$ be its rank in the WSR test
of that value. Then, $rank(d_{bi}^{\max})$  is the greatest value of $W^+$ and for any value $v$ such that $rank(v) > rank(d_{bi}^{\max})$ we have $time(S_b,j) - time(S_i,j) < 0$ and $v > d_{bi}^{\max}$. 
\end{pte}
{\bf proof:} By definition of the ranks and since $d_{bi}^{\max}$ is the largest positive value of \\
$time(S_b,j) - time(S_i,j)$ then it has the largest rank in $W^+$, thus any value having an absolute value greater than $d_{bi}^{\max}$ is negative and has a greater rank $\odot$

\begin{pte}
Suppose that for any subproblem $j$ the timeout for $j$ is set for $S_i$ to a value $to(j) > d_{bi}^{\max} + time(S_b,j)$ and let $W^+$ be the sum computed with these timeouts. Then, for any value of timeout greater than $to(j)$ the value of $W^+$ remains unchanged.   
\end{pte}
{\bf proof:} If the timeout is set to $to(j) > d_{bi}^{\max} + time(S_b,j)$ then for any $j$ reaching the timeout $|time(S_b)-time(S_i)| > d_{bi}^{\max}$. From Property \ref{d12max} the increase of $to(j)$ will not change the rank of the elements of $W^+$ so the property holds $\odot$.\\

So, for each subproblem $j$ such that $S_i$ has been stopped by a limit which is less than $d_{bi}^{\max} + time(S_b,j)$, we solve again this subproblem with $S_i$ within the time limit defined by  $d_{bi}^{\max} + time(S_b,j)+1$. Then, our deduction are statistically valid.

At the end, it is possible that we cannot deduce that some strategies are statistically different. However, this means that they should lead to equivalent solving time for the whole problem, so we can select any of them. in this case, we select the strategy performing the best on the sample.

If we compare $s$ strategies with an initial timeout fixed to twice the time of the best strategy and if $tmax(S_b)$ denote the largest solving time of a subproblem of the sample by $S_b$ the best strategy, then the sum of the solving times for all the strategies for each problem in the sample is bounded by $s \times tmax(S_b)$.

\paragraph{Significance level of the results}

The significance level of the method is bounded by the product of the confidence intervals of each comparison. This means that for $k$ comparisons, each with a confidence interval of $99\%$, the overall result has a confidence interval of $0.99^{k-1}$. Fortunately we have only few strategies. For instance for $7$ strategies, this leads to a confidence level of $0.99^6= 94\%$. This is quite acceptable.

\section{Related work}

There has been a significant amount of work to automatically select or adapt the search strategy. Some successes have been obtained by running some algorithms in parallel in CP \cite{gomes01} and in SAT \cite{hamadi09}. Offline and online machine learning based methods are popular. Offline methods select automatically the strategy among a set of available strategies. They perform a learning phase on a training set of instances. They haven been initially proposed for SAT \cite{xu08} and then for CSPs \cite{omahony08}. Y. Hamadi \cite{hamadi13} wrote a book on this subject and he proposed two methods: continuous search which aims at finding the best strategy for solving a given problem and autonomous search which aims at finding the best strategy in general. These methods are based on machine learning techniques. On the other hand, online methods have been considered. Epstein et al. \cite{epstein02}  proposed Adaptive Constraint Engine (ACE), a method which gathers the decision made by several strategies and proceed to a vote in order to decide which one will be applied for the next decision. Gagliolo and Schmidhuber \cite{gagliolo06} allocate times to each algorithms by using a multi-armed bandit algorithm whose decisions is based on the previous computations. Arbelaez et al. \cite{arbelaez09} apply Support Vector
Machines to the problem of automatically adapt the search strategy of a CP solver in order
to more efficiently solve a given instance. Loth et al. \cite{loth13} define the best strategy during the search by using a multi-armed bandit approaches combined to Monte Carlo Tree Search. 
For a good introduction to Algorithm selection we encourage the reader to refer to \cite{gagliolo06} and \cite{kotthoff16}.


\section{Experiments}

All the experiments have been run in parallel on a parallel machine. The scaling of the EPS method does not depend on the problem solved, so it is the same for all the variable-value strategies. Therefore, for each strategy we have used the sum of the time spent on each core allocated to this strategy as a measure of the time required by the strategy. The best of these times correspond to the value we want to minimize, thus our experiments are based on these times.

\paragraph{Machines}
All the experiments have been made on a Dell machine having  four   E7-4870 Intel processors, each having 10 cores with 256 GB of memory and running under Scientific Linux.

\paragraph{Solver}
We implemented our method on the top of Gecode 4.2 (http://www.gecode.org/).
 
\paragraph{Considered strategies}

After some experiments we selected 7 candidate strategies. Each strategy is dynamic:
\begin{itemize}
\item{FF} implements the first fail principle by selecting the variable with the minimum domain size \cite{haralick80};
\item{Act} selects the variable with the maximum of activity \footnote{Roughly the activity is defined by the number of times the variables has been introduced in the propagation queue. The activity is increased at most by one for each decision.} \cite{michel12};
\item{Wdegm} selects the variable with the maximum weighted degree \footnote{The weighted degree of a variable is defined by a a counter associated with it. Each time a constraint fails, the counter of each variable involved in the constraint  is increased by one.} \cite{boussemart04};
\item{WdegM}  same as above excepted that the value is selected differently;  
\item{MRegret} selects the variable for which the difference between the largest and second-largest value still in the domain is maximum \cite{gecode}.
\item{MostC} selects the most constrained variable. 
\item{D/Wdeg} selects the variable for which the ratio of the size of its domain by its weighted degree is minimum \cite{boussemart04,boussemart04b}.
\end{itemize}

After selecting the variable all strategies but Wdegm, assign to it the minimum value of its domain. WdegM assigns to it the maximum value of its domain.
We did not consider impact based strategy \cite{refalo04} because this strategy is not implemented in Gecode.

\paragraph{Benchmarks instances}

We present the most representative results that we obtained. Problems come from the CSPLib, the minizinc challenge \cite{MiniZInc2012} or the Hakank's constraint programming blog \cite{hakank}.

For satisfaction problems we search for all solutions and we consider the following problems:
all-I: All intervall series 14; Costa: Costa Array 13;
Filo: Filomino 13;
Lams 9;
Qgrp: Quasi group 7;
Msplt: Market split s5-08;
Sched: sport scheduling 12;
Tank: tank attack puzzle 7;
Gol: Golomb 12;
Perm: Permutation 12.

For optimization problems, we search for the optimal solution and we prove the optimality. Results are given for the following problems:
Crew; Dud: dudney thea; Java: java routing trip 6-3;
mario; mario medium 3; Fback: minimum feed back; matching problem  
Money: money change 27;
War: War Peace 8;
Sugi: Sugiyama 7 7;

\paragraph{Sampling}

The initial problem is decomposed into 16635 subproblems from which we randomly select 100 subproblems.

\subsection{Main results}

PSS denotes the Parallel Strategies Selection that we propose.

Times are expressed in minutes and correspond to the sum of the times spent by all the cores. 
Bold times indicate the best strategy for the considered problem.

\setlength{\tabcolsep}{5pt} 
\begin{center}
  \begin{tabular}{  | *{9}{c|}}
\hline
& FF & Act & Wdegm & WdegM & MRegret & MostC & D/Wdeg & PSS \\
\hline
All-I & 26.3	& 210& 55.1	& 54.4	&31.6	&26.1 & {\bf 0.8} & 0.9\\
Costa & {\bf 46.2}	& 365 &	78.2 &	153 &	213&	41.7& 96.9 & 49.2 \\
Filo & 427& 160	&	{\bf 12.0}&	78.2&	335&	654&	23.5 &12.4\\
Lams & {\bf 58.6}&	802&	416.3&	319.2&	49.9&	48.7&	1301 & 62.0\\
Qgrp & 36.7&	41.0&	367&	877&	4.6&	3.3&	{\bf 2.8}& 3.0\\
Msplt & 525&	1035&	616&	620&	526&	{\bf 492}& 703&	515\\
Sprt & 55.8&	265&	124&	116&	73.0&	36.6& {\bf 14.9} &	15.4\\
Tank & 29.6&	1091&	27K&	47K&	40.6&	13K& {\bf 3.8}&	4.1\\
Gol & 341&	295&	543&	455&183&	334&{\bf 168} &	176\\
Perm & 234&	177&	159&	201&	121&	331&	{\bf 27.3}&28.1\\
\hline
\end{tabular}
\end{center}

In term of ratio with respect to the best time (i.e., each time is divided by the best time), we obtain the following table which clearly shows the strong disparities between strategies and that the performance of PSS is close to the one of the best strategy for each problem. We use the following notation: $\overline{x}$ is the mean and geo $\overline{x}$ is the geometric mean.

\begin{center}
  \begin{tabular}{  | *{9}{c|}}
\hline
& FF & Act & Wdegm & WdegM & MRegret & MostC & D/Wdeg & PSS \\
\hline
All-I & 32 &	254	& 67	& 66 &	38.4&	31.7&	1 & 1.06\\
Costa & 1.1	&8.8	&1.9	&3.7	&5.1	&1.0	&2.3 & 1.06\\
Filo & 35	&1.0	&13	&6.5	&27	&54	&1.96 & 1.04\\
Lams & 1.2	&16.5	&8.6	&6.6	&1.0	&1.0	& 26.8 & 1.06\\
Qgrp & 13.1	&14.7	&131&	314	&1.6	&1.2& 1.0 &	1.06\\
Msplt & 1.1	&2.1	&1.3&	1.3	&1.1	&1.0& 1.43 &	1.04\\
Sprt & 3.7	&17.8 &8.4	&7.8	&4.9	&2.5	&1.0 &1.03\\
Tank & 7.9	&291	&7408	&12625	&10.8	&3576	&1.0 &1.07\\
Gol & 2.0 &	1.8	&3.2	&2.7	&1.1	&2.0	&1.0&1.05\\
Perm & 8.6	&6.5	&5.8	&7.4	&4.4	&12.1	&1.0&1.03\\
\hline
geo $\overline{x}$ & 5.1	& 12.1 & 17.6	&19.6	&4.4	&7.3	&1.7	&1.05\\
$\overline{x}$  & 10.6&61.5	&765	&1304	&9.7	&368.3 & 3.8	&1.05\\
\hline
\end{tabular}
\end{center}


For optimization problems we obtain the following results:
\begin{center}
  \begin{tabular}{  | *{9}{c|}}
\hline
& FF & Act & Wdegm & WdegM & MRegret & MostC & D/Wdeg & PSS \\
\hline
Crew	& 64	&258	&85	&91	&68	&{\bf 58}	&74 & 61\\
Dud &{\bf 15}&	34	&40	&32	&37	&17	&16.1 & 16.3\\
Java &24	&35	&41	&35	&{\bf 21}	&24	&108 & 22.7\\
Mario	&{\bf 4.2}	&45.9	&18.7	&9.8	& 7.4	& 5.8	& 5.9 & 4.6\\
Fback	&{\bf 126}	&281	&379	&436	&128	&131	&127 & 133\\
Money &	{\bf 0.6}	&0.9	&0.9	&0.6	&0.8	&0.6	&0.6 &0.6\\
War &185	&259	&232	&250	&211	&{\bf 54}	&176 & 56.4\\
Sugi &	504	&154&113&111&381	&504& {\bf 29.1} & 30.1\\
\hline
\end{tabular}
\end{center}

We can also expressed them in term of ratios w.r.t. the best time in order to see the relative differences between strategies:

\begin{center}
  \begin{tabular}{  | *{9}{c|}}
\hline
& FF & Act & Wdegm & WdegM & MRegret & MostC & D/Wdeg & PSS \\
\hline
Crew	& 1.10&	4.44&	1.46&	1.57&	1.17&	1.00& 1.21 & 1.05\\
Dud &	1.00&	2.23&	2.60&	2.06&	2.39&	1.12	& 1.07 & 1.07\\
Java &	1.12&	1.62&	1.91&	1.64&	1.00&	1.10& 5.14 & 	1.06\\
Mario&	1.0&	10.9&	4.43&	2.33&	1.75&	1.37&	1.40 & 1.09\\
Fback&	1.00&	2.22&	3.00&	3.45&	1.02&	1.04& 1.01 &	1.05\\
money& 	1.00&	1.57&	1.57&	1.09&	1.35&	1.12&1.05 & 	1.05\\
War & 3.45	&4.82&	4.32&	4.65&	3.92&	1.00& 3.26 &	1.05\\
Sugi & 17.3	&5.30&	3.90	&3.80&	13.0	&17.3&	1.00 & 1.05\\
\hline
geo $\overline{x}$ & 1.71	&3.35	&2.67	&2.32	&2.01	& 1.56	& 1.54 & 1.06\\
$\overline{x}$  & 3.38	& 4.14	& 2.90	& 2.57	& 3.21	& 3.13	& 1.88 & 1.06\\
\hline
\end{tabular}
\end{center}

Once again our method gives good results.
Note that for all problems the Wilcoxon signed rank test was able to eliminate all strategies against the best one.

\subsection{Comparison with Multi Armed Bandit (MAB) Approach}

The Multi-Armed Bandit selector is based on a model defined on a set of $k$ arms, one for each strategy, and a set of rewards $R_i(j)$, where $R_i(j)$ is the reward delivered when an arm $i$ has been chosen at time $j$. A reward reflects the performance of choosing that arm. 
The idea is to select for each subproblem a strategy (i.e., an arm) and then to solve the subproblem with this strategy. This will give us a reward inversely related to the solving time. The next selection is based on the sequence of the previous trials. We propose to use the UCB1 policy defined in \cite{auer02}, which selects the arm $i$ that maximizes 
$p(i)=\overline{R}_i + \sqrt{\frac{2ln m}{m_i}}$, where $m$ is the current number of selection, $m_i$ the number of times $i$ has been selected and $\overline{R}_i$ is the mean of the past rewards of the $i$ arm. 
This policy prefer the most rewarded strategy but also bias the selection toward less frequently selected strategies (this bias factor increases along the iterations).
The main difficulty is the definition of the reward function. We adapt the one of Gagliolo and Schmidhuber \cite{gagliolo06} which is designed for resource allocation and defined by: 
$\frac{\ln(t_{max}) - \ln(t_i)}{\ln(t_{max}) - \ln(t_{min})}$, where $t_{max}$ and $t_{min}$ are respectively the maximum and minimum solving time and $t_i$ is the time for solving problem $i$. Experimentally, we obtained the best results by defining $t_{max}=10\mu$ and $t_{min}=\mu/10$ where $\mu$ is the mean of the solving times. With such values we accept some variations and degenerated cases (i.e., very bad solving times) will give only negative rewards. We denote by MAB this method. Here is the comparison with PSS:


\begin{center}
  \begin{tabular}{  | *{5}{c|}}
\hline
 & \multicolumn{2}{c|}{time} & \multicolumn{2}{c|}{ratio w.r.t. best}\\ 
 \cline{1-5}
& PSS & MAB & PSS & MAB\\
\hline
All-I & 0.9	& 2.0 & 1.06 & 1.14 \\
Costa &49.2	&65.4 & 1.06 & 1.41\\
Filo &12.4	&36.8 & 1.04 & 3.08\\
Lams &62.0	&102 & 1.06 & 1.73\\
Qgrp &3.0	&7.8 & 1.06 & 2.77\\
Mspl &515	&548 & 1.04 & 1.11\\
Sprt &15.4&19.8 & 1.03 & 1.32\\
Tank &4.1	&12.0 & 1.07 & 3.13\\
Gol &176	& 243 & 1.05 & 1.45\\
Perm & 28.1	&31.4 & 1.03 & 1.15\\
\hline
geo $\overline{x}$ & & & 1.05 & 1.68\\
$\overline{x}$ && & 1.05 & 1.83\\
\hline
\end{tabular}
\end{center}
The results obtained with PSS are better than with  MAB. In addition PSS is more robust. These experiments show that applying the reasoning on subproblems coming from the instance to solve is certainly a good idea.


\subsection{Comparison with Portfolio}

PSS needs 1172 minutes for solving all the problems. The Portfolio-x4 method runs in parallel the four best strategies. It requires 3959 minutes which is not competitive with our method.



We also tried to combine our approach with a portfolio approach.
PSS-pfolio2 is the PSS method for which we run in parallel the two best estimated strategies when the difference between them is small.
The following results show that it is never interesting to run some strategies in parallel.

\begin{center}
  \begin{tabular}{  | *{7}{c|}}
\hline
& All-I & Costa & Lams & Qgrp & Msplt & Perm\\
\hline 
PSS & 0.9 & 49.2 	& 62.0 & 3.0& 515& 28.1\\
\hline
PSS-pfolio2 &1.6&	94.0&	114& 5.4&917&37.9\\
\hline
\end{tabular}
\end{center}

\subsection{Timeout, Sample size and Simple impact}

The timeout (TO) may have a huge impact on the selection time as shown by the following table, where``without TO'' means that we do not stop any strategy when solving a subproblem. 
\begin{center}
  \begin{tabular}{  | *{3}{c|}}
\hline
& with TO & without TO\\
\hline
All-I & 0.1	& 4.9\\
Costa &3.0	&16.1\\
Filo &0.4	&17.5\\
Lams &3.4	&9.6\\
Qgrp &0.2	&2.9\\
Msplt &22.9	&82.8\\
Sprt &0.5	&7.8\\
Tank &0.3	&136\\
Gol &7.9	& 38.0\\
Perm &0.8	&5.5\\
\hline
\end{tabular}
\end{center}


We also made some experiments with a sample size equals to $30$ instead of $100$. We do not observe any difference for the selected strategy. The best strategy is selected for all problems.

\section{Conclusion}

The Embarrassingly Parallel Search method solves a problem by decomposing it into subproblems. In order to select the best variable-value strategy to solve a problem, we propose to use a part of these subproblems and compare some strategies on them. Then, we select the most promising one by using the Wilcoxon signed rank test. This method, PSS, is simple and does not require a lot of computations. It can easily be used in practice because the time allocated to the strategy selection is under control. Some comparisons with other portfolio approaches show the advantage of our method. We also give a model based on  the Multi Armed Bandit algorithm which gives interesting results although inferior and less robust than those of PSS. At last, it appears that it is better to select only one variable-value strategy than running several in parallel, even if we make some mistakes sometimes.
 
 \pagebreak
 
\bibliographystyle{plain}
\bibliography{jcr}

\end{document}